\documentclass[letterpaper]{article} 
\usepackage{aaai24}  
\usepackage{times}  
\usepackage{helvet}  
\usepackage{courier}  
\usepackage[hyphens]{url}  
\usepackage{graphicx} 
\usepackage{natbib}  
\usepackage{caption} 
\usepackage{algorithm}
\usepackage{algorithmic}
\usepackage{booktabs} 
\usepackage{multirow} 
\usepackage{newfloat}
\usepackage{amsmath}
\usepackage{listings}

\DeclareCaptionStyle{ruled}{labelfont=normalfont,labelsep=colon,strut=off} 
\floatstyle{ruled}
\newfloat{listing}{tb}{lst}{}
\floatname{listing}{Listing}
\usepackage[textsize=scriptsize]{todonotes}

\title{Detecting and Preventing Hallucinations in \\ Large Vision Language Models}
\author{
  Anisha Gunjal\equalcontrib,
  Jihan Yin\equalcontrib ,
  Erhan Bas\thanks{Work done at ScaleAI} \\ 
}
\affiliations{
    Scale AI\\
    {anishagunjal@utexas.edu, 
    jihan\_yin@berkeley.edu,
    erhan.bas@gehealthcare.com} 

}

\usepackage{bibentry}

\begin{document}

\maketitle

\begin{abstract}
Instruction tuned Large Vision Language Models (LVLMs) have significantly advanced in generalizing across a diverse set of multi-modal tasks, especially for Visual Question Answering (VQA). However, generating detailed responses that are visually grounded is still a challenging task for these models. We find that even the current state-of-the-art LVLMs (InstructBLIP) still contain a staggering 30 percent of the hallucinatory text in the form of non-existent objects, unfaithful descriptions, and inaccurate relationships. To address this, we introduce \textbf{M-HalDetect}, a \textbf{M}ultimodal \textbf{Hal}lucination \textbf{Detect}ion Dataset that can be used to train and benchmark models for hallucination detection and prevention. M-HalDetect consists of 16k fine-grained annotations on VQA examples, making it the first comprehensive multi-modal hallucination detection dataset for detailed image descriptions. Unlike previous work that only consider object hallucination, we additionally annotate both entity descriptions and relationships that are unfaithful. To demonstrate the potential of this dataset for hallucination prevention, we optimize InstructBLIP through our novel Fine-grained Direct Preference Optimization~(FDPO). We also train fine-grained multi-modal reward models from InstructBLIP and evaluate their effectiveness with best-of-n rejection sampling~(RS). We perform human evaluation on both FDPO and rejection sampling, and find that they reduce hallucination rates in InstructBLIP by 41\% and 55\% respectively. We also find that our reward model generalizes to other multi-modal models, reducing hallucinations in LLaVA and mPLUG-OWL by 15\% and 57\% respectively, and has strong correlation with human evaluated accuracy scores. The dataset is available at \url{https://github.com/hendryx-scale/mhal-detect}.
\end{abstract}

\section{Introduction}

\begin{figure*}
    \centering
    \includegraphics[scale=0.65,trim=75mm 105mm 70mm 40mm]{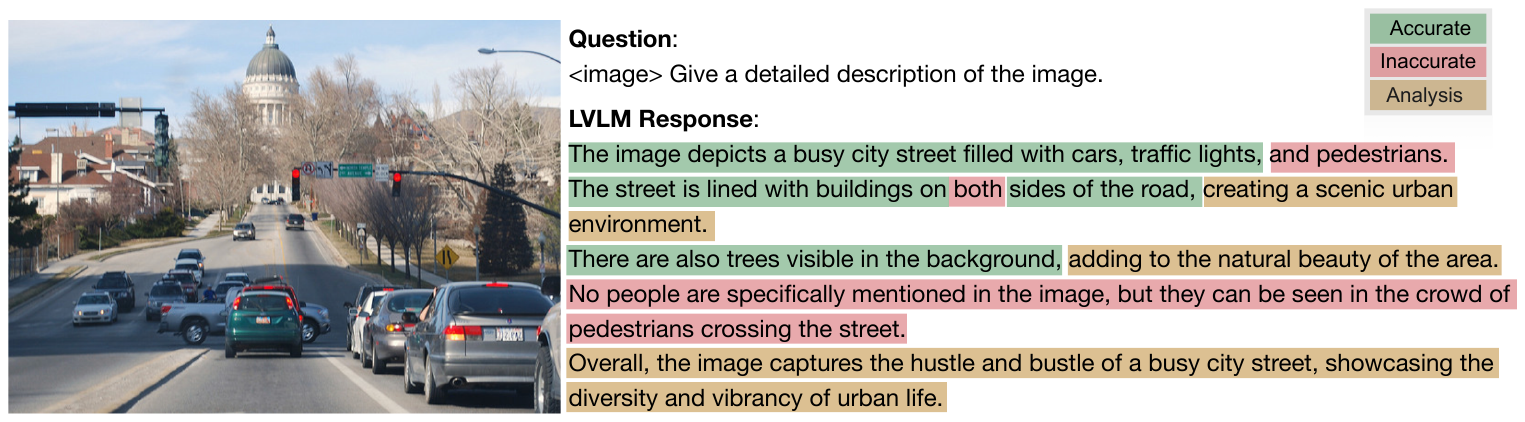}
    \caption{Example Annotation from the M-HalDetect Dataset. The sub-sentences of text generated by multi-modal LM are tagged into categories:\textit{ Accurate, Inaccurate}, and \textit{Analysis}. }
    \label{fig:data-example}
\end{figure*}

Large language models (LLMs) have transformed the AI landscape in recent years, scaling their training data to trillions of tokens and their parameter count to hundreds of billions \cite{brown2020gpt, achiam2023gpt, touvron2023llama}. This has unlocked powerful emergent behaviors, and seen widespread adoption through the use of chat agents such as ChatGPT. Recently, advances in multi-modal models have seen adoption around grafting visual backbones onto pre-trained large language models, resulting in LVLMs \cite{liu2023visual, instructblip, ye2023mplug}. While this has led to strides in overall VQA performance, it brings along the same challenges that plague these LLMs - a significant one being the propensity to generate hallucinations.

In language models, hallucinations occur when the model produces inaccurate or misleading factual information that cannot be supported by existing knowledge stores \cite{ji2023survey, bang2023multitask}. In the context of VQA for LVLMs, hallucinations can manifest as responses containing references or descriptions of the input image that are incorrect \cite{li2023evaluating}. It is essential to address and mitigate these hallucinations to enhance the reliability and accuracy of multi-modal models in real-life usecases. However, these multi-modal hallucinations are hard to programatically detect and often requires human supervision, which can be costly.

To facilitate automatic hallucination detection, we first build a diverse human-labeled dataset using VQA responses from InstructBLIP, as seen in Figure \ref{fig:data-example}. We then train multiple reward models of various densities (sentence-level, sub-sentence level) on this dataset for hallucination detection. An effective way to use these reward models to reduce hallucinations is to use them to generate rewards in a reinforcement learning setup \cite{ziegler2019finetune, stiennon2020learning, nakano2021webgpt}, although the resulting final model can only be as effective as the original reward model used \cite{bai2022training}. Therefore, in this paper we focus on measuring the quality of these reward models, exploring classification metrics and using best-of-n rejection sampling as an approximation of the system's performance. Similar to \cite{rafailov2023direct}, we also directly optimize InstructBLIP with fine-grained Direct Preference Optimization (FDPO), a novel variation of DPO in which we leverage fine grained annotation information from individual examples, rather than collecting relative preference signals from pairs of texts. Both methods show significant success in reducing hallucination rates from InstructBLIP, and furthermore, rejection sampling with our reward models reduces hallucination rates in other multi-modal models as well - LLaVA \cite{liu2023visual} and mPLUG-OWL \cite{ye2023mplug}. 

\pagebreak
Our main contributions are as follows:

\begin{enumerate}
    \item We create and release M-HalDetect, a new hallucination detection dataset focused on fine-grained annotations at a sub-sentence level over detailed image descriptions.
    \item We show that InstructBLIP can be optimized using Fine-grained DPO (FDPO) using the M-HalDetect dataset to reduce hallucination rates by 41\%.
    \item We show that reward models trained on this dataset can reduce hallucination rates by 55\% in InstructBLIP with best-of-64 rejection sampling. The reward model generalizes to other LVLMs, reducing hallucination rates in LLaVA and mPLUG-OWL by 15\% and 57\% respectively with best-of-16 sampling.
    \item We show that our reward model is an effective evaluator of hallucination rates, giving scores aligned with human ratings.
\end{enumerate}


\section{Related Work}

Large Vision Language Models (LVLMs) have seen performative advancements in tasks such as generating text from images\cite{li2023large} and multi-modal in-context learning\cite{alayrac2022flamingo}. Recent work has focused on utilizing instruction tuning techniques to enhance the zero-shot performance of instruction-aware LVLMs across different vision-language tasks\cite{liu2023visual,instructblip}. These approaches utilize GPT-4 to generate multi-modal instruction tuning datasets\cite{liu2023visual} where the image context is provided to GPT-4 through symbolic representations of the image such as captions and object bounding boxes. Others combine datasets across various multi-modal tasks\cite{instructblip} with hand-crafted instructions, a method that has found success in training traditional LLMs\cite{wei2021flan}. This achieves state-of-the-art performance in a variety of multi-modal tasks such as visual and video QA, image captioning and classification.

Nevertheless, a significant challenge associated with LVLMs has emerged: preventing hallucinations when generating textual output. It is essential to address and mitigate these hallucinations to enhance the reliability and accuracy of LVLMs in production use cases.

\paragraph{Hallucination Analysis in LVLMs} In \cite{li2023evaluating}, the evaluation metric "POPE" is proposed to evaluate hallucinations in LVLMs by polling questions about generated text. They observed that current state-of-the-art LVLM (InstructBLIP) has the lowest object hallucination rates among recent LVLMs. Another relevant contribution by Liu et al. \cite{liu2023aligning} is the introduction of the LRV dataset. This dataset contains positive and negative instructions specifically designed to enhance the robustness of LVLMs against hallucination and inconsistent text generation. Furthermore, they proposed a method called GAVIE, which leverages GPT-4 to assist in evaluating preferred answer generations. 

These studies collectively contribute to the understanding and mitigation of hallucination-related challenges in LVLMs, by providing evaluation metrics, datasets, and evaluation methods that enhance the reliability and consistency of text generation in multi-modal models. Our work extends the scope of the previous works by not only considering hallucinations on the presence of objects, but also on descriptions of objects such as relative positioning or attributes. We also consider hallucinations on complex object reasoning.

\paragraph{Aligning to Human Preferences}
Despite having strong zero-shot performance on classical language benchmark datasets, pre-trained LLMs still struggle to produce detailed generations on par with those written by real humans. Supervised fine-tuning on demonstration data written by humans is not enough, where recent works have focused on using Reinforcement Learning with Human Feedback (RLHF) to address this problem \cite{stiennon2020learning,touvron2023llama,ouyang2022training, achiam2023gpt}.

RLHF typically uses Proximal Policy Optimization \cite{schulman2017ppo}, to optimize a policy model with rewards from a reward model. This reward model is typically trained on preference pairs of same-prompt generations, often sourced from the base policy model. This preference is usually given by humans, though attempts have been made to use more traditional metrics such as BLEU \cite{papineni2002bleu} and ROUGE \cite{ganesan2018rouge} as proxies. Using human preferences is more effective in aligning LLMs to human preferences \cite{stiennon2020learning}, though sees mixed results in hallucination prevention. Ouyang et al. \cite{ouyang2022training} found that RLHF helps smaller (6B) language models reduce their hallucination rate, while having the opposite effect on larger models (175B). In this paper, we will focus on relatively smaller multi-modal models (7B) that can be more accessible to end users.

DPO has emerged recently as a viable alternative to RLHF for preference alignment, optimizing the policy model directly without needing to train a reward model and sample rewards through reinforcement learning \cite{rafailov2023direct}. It has shown comparable performances with RLHF in summarization and chatbot usecases on language models, and  maintains strong performance in higher temperature sampling. At the same time, it avoids the unstable and brittle process of training models with RL \cite{engstrom2020implementation}.

\paragraph{Fine-grained Preferences}

A limitation of both RLHF and DPO is their lack of fine-grained interpretability regarding what makes one generation more preferred than the other. Recent research has made significant progress in leveraging fine-grained user preferences to improve the performance and interpretability of reward models. For example, Wu et al. \cite{wu2023fine} utilize fine-grained human feedback to train multiple reward models at different density levels. These reward models covered passage level preferences as in the traditional RLHF setting, but also sentence level and sub-sentence level preferences in the form of error identification. \cite{lightman2023let} employs process supervision, providing human feedback on individual steps for more robust rewards.

To extend this fine-grained feedback mechanism into the multi-modal domain, we introduce a new dataset for multi-modal hallucination detection. Our dataset comprises of 4,000 images with 4 detailed descriptions each, for a total of 16,000 image description pairs, annotated at the sub-sentence level to indicate the accuracy of the generated descriptions. Similarly to \cite{wu2023fine}, we train sub-sentence and sentence level reward models on this dataset. We also modify the DPO loss to utilize fine-grained annotations.

\section{M-HalDetect : Multi-Modal Hallucination Detection Dataset}
\paragraph{Dataset Description} In this section, we introduce the M-HalDetect dataset that incorporates fine-grained annotations for identifying hallucinations in detailed image descriptions generated by LVLMs. The dataset comprises of image-description pairs sampled from 4,000 images taken from the \textit{val2014} split of the Common Objects in Context (COCO) dataset \cite{lin2014coco}. The dataset is divided into a training set with 3,200 images and a development set with 800 images. 

We choose to utilize the validation set of COCO to avoid potential training data regurgitation from LVLMs trained on the COCO training set. This is roughly 10\% of the original COCO validation set, leaving enough data untouched to not impact further validation too heavily.

To generate responses, we prompt InstructBLIP \cite{instructblip} with each image and a randomly selected question from a pool of instructions for describing an image. We initially reuse instructions from ones used in InstructBLIP's detailed image description training data, which were sourced from the LLaVA-150k \cite{liu2023visual} dataset. During initial analysis, we observed that doing so led to less diverse responses, potentially due to the influence of this dataset during training. To address this, we added in our own prompts to improve generation diversity.\footnote{Refer to \url{https://arxiv.org/abs/2308.06394} for details on dataset and diverse prompt generation, training, and inference analysis.}

We sample four responses using nucleus sampling from InstructBLIP with a temperature value set to 1.0. This creates 16k image-prompt-response triplets, split between 12800 samples in the \textit{train} split and 3200 samples in the \textit{val} split.

\paragraph{Dataset Categories} The annotation process involves categorizing different segments of each response into three categories: (i) Accurate, (ii) Inaccurate, and (iii) Analysis. We also include an Unsure category for ambiguous cases. We define the classes as follows:
\begin{itemize}
    \item \textbf{Accurate} Objects exist in the image, their descriptions are accurate according the image, and any described relationships can be accurately inferred from the image.
    \item \textbf{Inaccurate} Objects do not exist in the image or their descriptions are inaccurate. Furthermore, if the analysis about the image is not plausible, it is also marked as Inaccurate.
    \item \textbf{Analysis} Scene or object analysis including complex reasoning or interpretations about the image. These are portions of the data that are more subjective and not grounded visually within the image.
    \item \textbf{Unsure} This category is reserved as a last resort if annotators cannot make a judgment about the sentence segment into one of the above three categories.
\end{itemize}

We provide fine-grained annotations for these 3 categories on the detailed descriptions of images generated by the LVLM. The annotations are provided at sub-sentence level - i.e. one sentence can comprise of multiple segments from different classes, as seen in Figure \ref{fig:data-example}.

To make the annotation process user-friendly, we allow a leeway to the annotators to miss a few words in the annotations if there are too many segments in a sentence to be annotated. The unmarked words in a sentence are by default considered as "Accurate". In our analysis, we noticed that sometime annotators skip annotating punctuation, connector words, or introductory sub-sentences such as "The image features" (illustrated in Figure \ref{fig:data-example}). 

\paragraph{Dataset Collection} To collect the annotations, we employed Scale AI's RAPID\cite{scalerapid} labeling tool and involved 10 randomly selected human annotators. These annotators had to qualify by passing a training course with a minimum accuracy of 85\% on the example tasks to be selected for the final tagging task. The annotators are presented with an image and four responses about the image generated by InstructBLIP. Their task is to annotate segments of the sentence into one the categories. An example annotation task is illustrated in Figure \ref{fig:data-example}.

\section{Method}

\subsection{Multi-Modal Reward Model} 
We implement a multi-modal reward model for detecting the presence of hallucinations generated by LVLMs. Specifically, we reuse the InstructBLIP weights and architecture, swapping the final embedding layer with a classification head. We do this as initializing the reward model from the generative model weights improves training robustness and reward generalization in later RL \cite{zheng2023secrets}. InstructBLIP consists of an image encoder that extracts image features and a linear mapping layer that projects these features. These image feature are passed to an instruction-aware attention layer, the QFormer, that attends instructions over the projected image features. The QFormer outputs are passed to a frozen pretrained decoder as soft prompts, prefixed to the instruction. For this paper, we choose to use Vicuna \cite{vicuna2023} as the frozen decoder following the original InstructBLIP.

We train reward models at sentence level and sub-sentence level densities. For each image-text pair, we run one forward pass similar to \cite{lightman2023let}, and set target class labels at the token concluding each segment, masking out all other indices in the segment. We optimize with cross-entropy loss. We fine-tune the entire decoder and reward model head, while freezing the rest of the model. Ablations on model freezing, hyperparameters as well as details on training can be found in the extended version.

\label{sec:sent-level_rp}
\begin{figure}
    \centering
    \includegraphics[scale=0.5,trim=100mm 70mm 100mm 70mm]{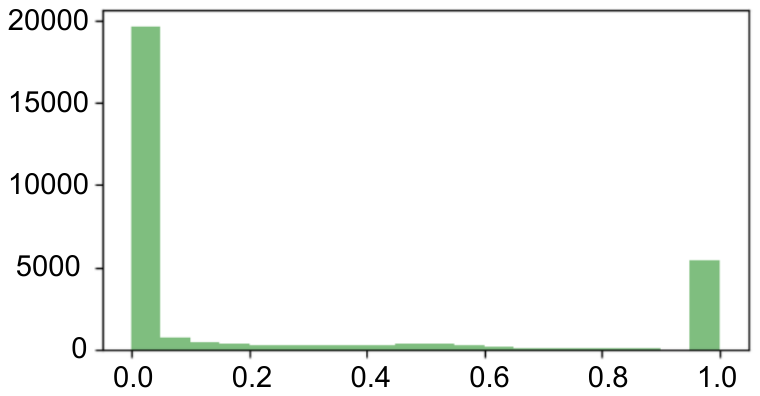}
    \caption{Label density histogram for the Inaccurate class. The x-axis represents the percentage of a sentence that is annotated as Inaccurate and the y-axis represents the frequency of such sentences in the dataset.}
    \label{fig:inaccurate-density}
\end{figure}
\subsection*{Sentence-level Reward Prediction}
We condense the labeled sub-sentence segments in M-HalDetect into sentence-level segments for a more structured reward format - this makes it more straightforward to run rejection sampling and train with RL, without worrying about localizing proper segments.  We identify these sentences using the Natural Language Toolkit\cite{bird2009natural}. For each sentence, if there is any segment that is inaccurate, we label the entire sentence as inaccurate. While this may introduce some noise when converting partially inaccurate sentences, we see in Figure \ref{fig:inaccurate-density} that the frequency of such sentences is low. Furthermore, if a sentence has a segment with the "unsure" category, we merge that sentence into the inaccurate class.  We experiment with two levels of label granularity with this dataset:
\begin{itemize}
    \item \textbf{Binary Classification}: Condense Analysis and Accurate classes into the Accurate class. In this setting we have two classes: \texttt{Accurate} and \texttt{Inaccurate} 
    \item \textbf{Ternary Classification}: In this setting, we have three classes: \texttt{Accurate}, \texttt{Inaccurate} and \texttt{Analysis}.
\end{itemize}

\subsection*{Segment-level Reward Prediction}
We also train a finer-grained reward model that make hallucination judgments on segments of sentences as opposed to entire sentences. This can provide less noisy signal when training on annotations, especially with longer compound sentences and hallucinations isolated to small portions of a sentence. We train on this data in a similar fashion to the sentence level rewards, by labeling the end token index of each span or segment of annotated text into its corresponding label. We then mask out every other index in the sequence. As a baseline, we assume perfect localization of the annotation segments as an upper bound for the performance of this method. Future works can consider training a segment localization model in parallel with the reward model, to detect when hallucinations start and end. Since we do not do this, we cannot use this reward model for rejection sampling, and evaluate purely on classification metrics over the test set. Similar to sentence-level reward prediction baselines, we also experiment with the binary and ternary variants of the segment-level reward prediction models. 

\begin{figure}
    \centering
    \includegraphics[scale=0.35,trim=30mm 50mm 20mm 40mm]{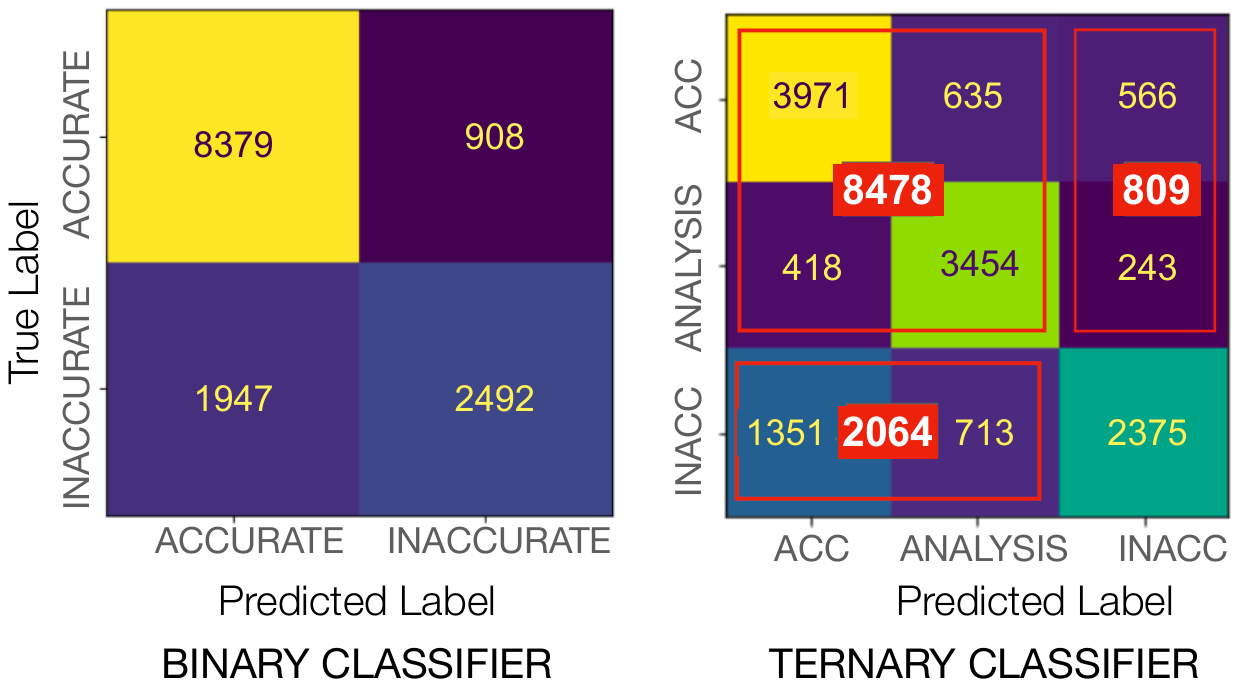}
    \caption{Confusion Matrix comparison between Binary and Ternary Classifiers. The right plot represents the binary classifier labels derived from the ternary classifier by merging the Accurate and Analysis classes.}
    \label{fig:cm_dt_decoder}
\end{figure}

\begin{figure*}
    \centering
    \includegraphics[scale=0.82,trim=285mm 85mm 300mm 80mm]{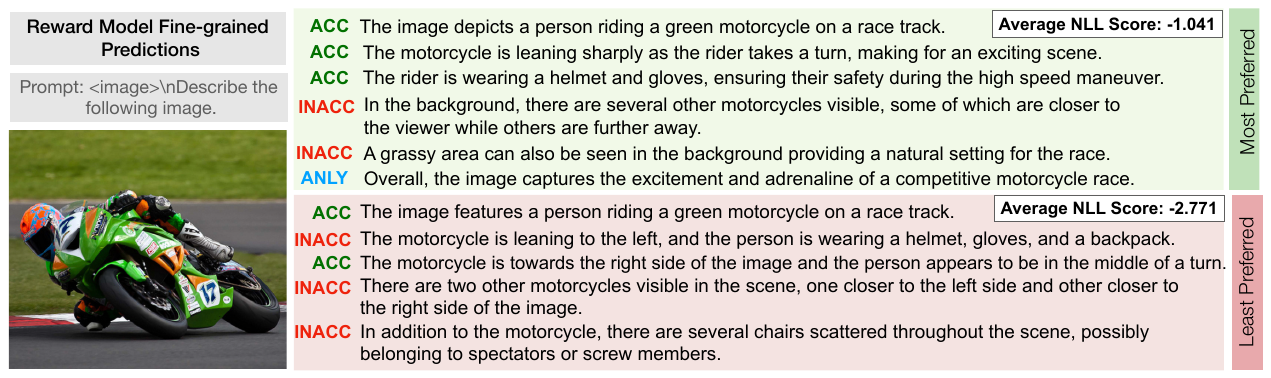}

    \caption{Rejection sampling examples with ternary reward model labels per sentence. Score for each response is computed using the average negative log-probability per sentence of a hallucination.}
    \label{fig:rejection_sampling}
\end{figure*}

\subsection*{Rejection Sampling~(RS)}
We use the trained reward models to perform rejection sampling on the generations of InstructBLIP to promote selection of less hallucinatory responses. We do this on the passage level, computing reward scores for the whole generation at once. We calculate the reward score by averaging the non-hallucination negative log probabilities of each sentence. This represents the normalized negative log probability of the entire passage containing no hallucinations. We compute rejection sampling in a best-of-n and worst-of-n setting, for $n=16,64$, to study the ability of the reward model in selecting the best generations from InstructBLIP, and the variance in quality between generations.

As we train two types of sentence level reward models (binary and ternary, including the analysis class), we experiment with using both models for reward scoring. We found in our initial experiments that although the binary reward model is able to penalize hallucinations with low scores, it tends to give very high scores towards the analysis class. We theorize that it is much easier to detect non-hallucinogenic analysis over factual descriptions, and as a result the binary reward model scores are biased towards generations that contain more subjective analysis rather than objective descriptions. This is less of a problem with the ternary reward model, as analysis has been split into its own class. As we will discuss in the results, the ternary model's functionaltiy is a superset of the binary model. For these reasons, we choose to use the ternary reward model for rejection sampling moving forward.

To study our the robustness of our reward model and our dataset, we conduct rejection sampling on generations from other LVLMs, namely LLaVA and mPLUG-OWL. For these experiments, we reuse the reward model initialized from InstructBLIP.

\subsection{Fine-grained Direct Preference Optimization}
While we train a reward model to show the potential of optimizing against hallucinations with RL, we also directly optimize InstructBLIP using FDPO to reduce hallucinations. 

Since M-HalDetect does not contain the traditional preference pairs used in DPO and RLHF, we explicitly segment each generation into sequences of preferred, dispreferred, and neutral chunks. We then reuse the DPO loss in increasing the likelihoods of preferred chunks while decreasing the likelihood of dispreferred chunks, each regularized by the original likelihood from the base model for the corresponding chunk, while neutral chunks are ignored. Similar to \cite{wu2023fine}, this should give stronger signal during training in reducing hallucinatory generations as compared to using pairs of likelihoods over entire generations.

Recall the loss used in DPO, with $\pi_{ref}$ as the reference model, $\pi_{\theta}$ as the policy model, $x$ being the input, $y_{w}$ being the preferred generation, and $y_{l}$ being the dispreferred generation.


$$
\mathcal{L}_{\mathrm{DPO}}\left(\pi_\theta\pi_{\mathrm{ref}}\right) =-E_{\left(x, y_w, y_l\right) \sim \mathcal{D}}\left[\log \sigma\left( \Delta_r \right) \right] 
$$
$$
\Delta r = \beta \log \frac{\pi_\theta\left(y_w \mid x\right)}{\pi_{\mathrm{ref}}\left(y_w \mid x\right)} - \beta \log \frac{\pi_\theta\left(y_l \mid x\right)}{\pi_{\mathrm{ref}}\left(y_l \mid x\right)}
$$

Since we don't have preferences over pairs of generations, but spans of fine-grained preferences throughout each generation, our FDPO loss can be modeled as 

$$
L_{\mathrm{FDPO}}\left(\pi_\theta ; \pi_{\mathrm{ref}}\right)=-E_{\left(x, y, c\right) \sim \mathcal{D}}\left[\log \sigma\left(\beta k \right)\right]
$$
$$
k=
\begin{cases} 
    -r & c = 0 \\
    r & c = 1 \\
    -\infty & c > 1
   \end{cases},
\;\;\;\;\;\;
r = \log \frac{\pi_\theta\left(y \mid x\right)}{\pi_{\mathrm{ref}}\left(y \mid x\right)}
$$

with sample segments $x, y, c$ being drawn from the dataset. Here, $x$ is the entire input up until the start of the current segment, $y$ is the generated segment, and $c$ is the class of the current segment, with $c=1$ being the preferred class, $c=0$ being the dispreferred class, and all other classes being ignored. Since segments are non-overlapping, we can run a single forward pass for each sample to calculate the loss of all segments within the sample all at once.

This formulation allows us to categorize each class into positive, negative, or neutral signal, the latter of which will be ignored during training. We run ablations on including the analysis class as either a negative or neutral class when optimizing InstructBLIP with FDPO. We fine-tune only the QFormer and language head, keeping the rest of the model frozen. We use $\beta=0.5$ for all our FDPO experiments, and train for a maximum of 5 epochs with $lr=10^{-6}$, warmup ratio of $.03$, and a cosine scheduler. 

\begin{table*}[htbp]
    \centering
    \begin{tabular}{ccccc}
        \toprule
        Model & Type & Method & RM Score $\downarrow$ & Human Eval $\uparrow$ \\
        \midrule
        InstructBLIP & Baseline & Baseline (T=0) & 0.97 & 0.71 \\
        \midrule
        InstructBLIP & DPO & IA Finetune Qformer (T=0) & $\boldsymbol{0.48}$ & $\boldsymbol{0.83}$ \\
        InstructBLIP & DPO & IA Finetune Qformer (T=1) & 0.72 & 0.75 \\
        InstructBLIP & DPO & DA Finetune Qformer (T=0) & 0.85 & 0.70 \\
        InstructBLIP & DPO & DA Finetune Qformer (T=1) & 1.03 & 0.58 \\
        \midrule
        InstructBLIP & RS & Best of 64 & $\boldsymbol{0.26}$ & $\boldsymbol{0.87}$ \\
        InstructBLIP & RS & Worst of 64 & 1.76 & 0.53 \\
        InstructBLIP & RS & Best of 16 & 0.36  & 0.82 \\
        \midrule
        LLaVA & Baseline & Baseline (T=0) & 0.383 & 0.805 \\
        LLaVA & RS & Best of 16 & $\boldsymbol{0.159}$ & $\boldsymbol{0.834}$ \\
        \midrule
        mPLUG-OWL & Baseline & Baseline (T=0) &  1.26 & 0.476 \\
        mPLUG-OWL & RS & Best of 16 &  $\boldsymbol{0.595}$ & $\boldsymbol{0.707}$ \\
        \bottomrule
    \\
    \end{tabular}
    
    \caption{Results of reward model and human evaluation scores. The RM Score is the average negative log probability of the passage not containing hallucinations, while the human evaluation score is the percentage of content that was truthful. A perfect RM score would be 0, and a perfect human evaluation score would be 1.}
    \label{tab:humanevalres}
\end{table*}

\subsection{Evaluation} 
Recent works in multi-modal LLMs\cite{liu2023visual, liu2023aligning} sometimes use GPT-4 as a human proxy to qualitatively evaluate LM outputs. Specifically, GPT-4 is prompted to give a preference score to a LM generation, either as a stand-alone or compared against GPT-4's own generation. This metric enables automatic evaluation without depending on human evaluators.

However, this is plagued with systematic bias such as senstitivity to the ordering of responses \cite{wang2023large}. Furthermore, GPT-4's public API does not yet support image inputs. Recent multi-modal works instead pass image context in the form of captions and object bounding boxes. In several cases, this symbolic input cannot represent the image robustly and leads to incorrect evaluations. We performed a qualitative analysis on GPT-4's performance on LLaVA-150k's detail subset and noted that GPT-4 gave frequent inaccurate scores and explanations, failing to detect hallucinations while incorrectly penalizing correct generations. For this reason, we do not use GPT-4 for automatic evaluation of generation quality.

To combat these limitations, we use human evaluation to evaluate the hallucination rates of our rejection sampling and DPO generations. Following the same labeling instructions as the M-HalDetect, we annotate the generations into accurate, inaccurate, and analysis spans. For generations from our DPO model, we use temperature=1 and nucleus sampling. We apply this across 50 different images sourced from COCO's validation set, separate from the ones used in M-HalDetect, though we reuse instructions from the dataset.

A common trade-off between reducing hallucinations is a reduction in helpfulness. Consider, for example, a model that outputs nothing - it does not hallucinate, yet it is not helpful either. To avoid this potential bias in our evaluation, we choose to measure the hallucination rate as the number of inaccurate words divided by the number of total words, excluding analysis segments, to calculate what percentage of descriptive objective content contained hallucinations.

\section{Results}

\subsection{Reward Model Classification Metrics}
\begin{table}[htbp]
    \centering
    \begin{tabular}{cccc}
        \toprule
        \textbf{Type} & \textbf{Density} & \textbf{Accuracy} & 
        \textbf{F1 Score} \\
        \midrule
        Binary & Sentence Level & 79.2 & 78.37 \\
        Ternary & Sentence Level & 71.4 & 70.8 \\
        \midrule
        Binary & Segment Level & 83.92 & 83.22 \\
        Ternary & Segment Level & 77.2 & 76.93 \\
        \bottomrule \\
    \end{tabular}
    \caption{Baseline Reward Model Results}
    \label{tab:classification_res}
\end{table}

We evaluate the multi-modal reward models (sentence-level and segment-level) using the development split of the M-HalDetect Dataset. We report \textit{Accuracy} and \textit{F-1 Score} for each of the training strategies. All models are initialized with pre-trained InstructBLIP weights, and the results are reported in Table \ref{tab:classification_res}.

Although the binary version has higher accuracy and F1 than the ternary in both sentence and segment level applications, we see in Figure \ref{fig:cm_dt_decoder} that the ternary reward model actually performs about the same as the binary reward model, if we were to reduce from a ternary to a binary setting. The ternary model additionally learns to separate the Accurate and Analysis classes, and we use it for rejection sampling and reward scoring experiments moving forward.

\subsection{Human Evaluation}

Figure \ref{fig:rejection_sampling} illustrates an example of rejection sampling using fine-grained feedback from the reward model. The reward model can accurately flag hallucinatory sentences which incorrectly claims the presence of other motorcycles and chairs. Furthermore, it is also able to flag sentences that generate analysis about non-existent objects.

\begin{figure}
    \centering
    \includegraphics[scale=0.30,trim=50mm 30mm 50mm 10mm]{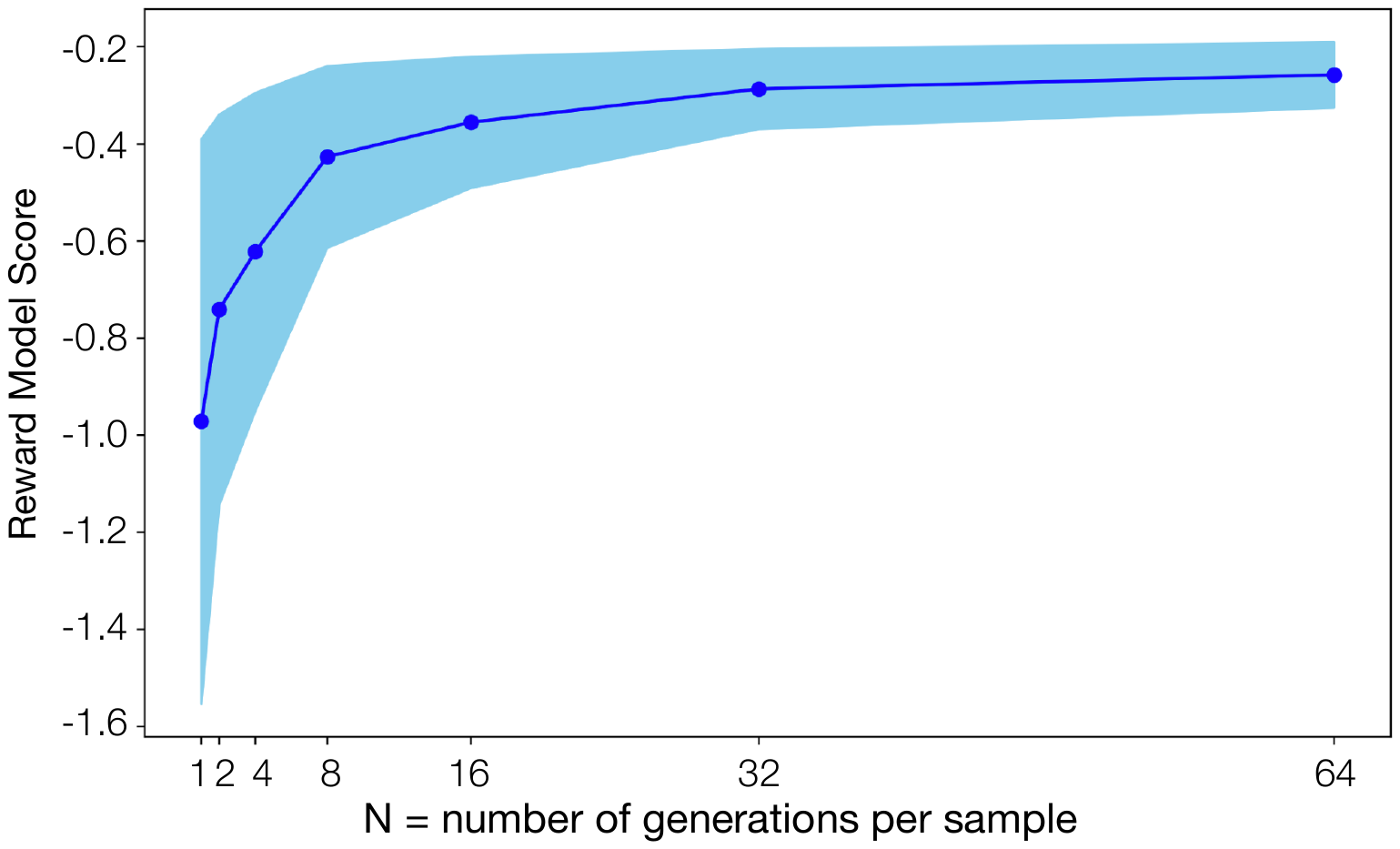}
    \caption{Reward model score means and variances as n increases in best-of-n rejection sampling. We see diminishing returns as we increase n.}
    \label{fig:rejection-sampling-distribution}
\end{figure}

We observe in Table \ref{tab:humanevalres} that rejection sampling significantly improves the factual rate of InstructBLIP's outputs. On the other hand, the worst generations of InstructBLIP can be extremely poor, with an almost 50\% hallucination rate! We can see from both the human eval results and our reward model scores in Figure \ref{fig:rejection-sampling-distribution} that we get exponentially diminishing returns as the sample size increases.

\paragraph{Rejection Sampling}

\begin{figure}
    \centering
    \includegraphics[scale=0.3]{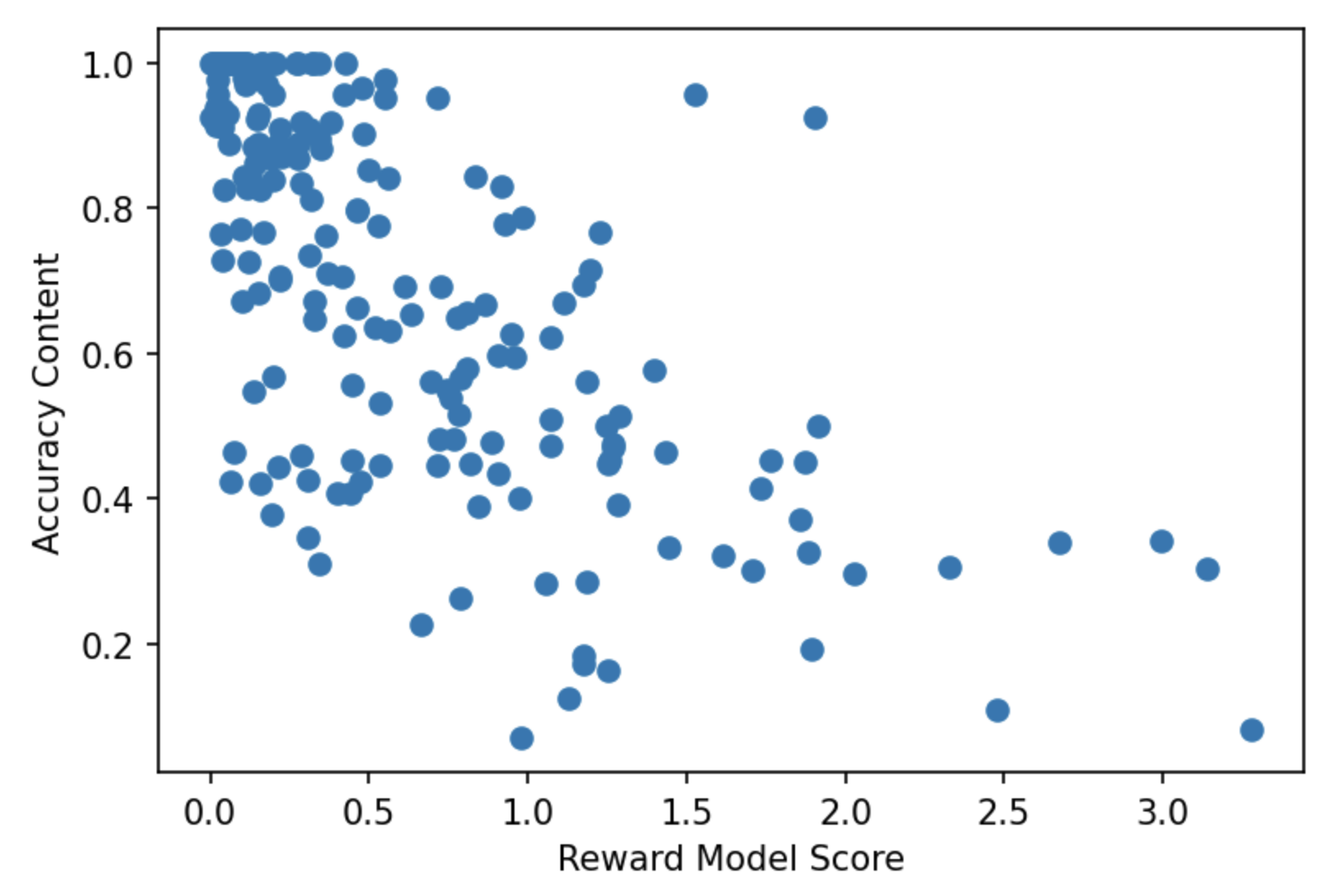}
    \caption{Human evaluation scores against reward scores for all human evaluated results.}
    \label{fig:reward_human_evaluation_correlation}
\end{figure}

We also see that rejection sampling with InstructBLIP manages to reduce hallucination rates for LLaVA and significantly for mPLUG-OWL. This shows that although M-HalDetect's image descriptions are sourced from InstructBLIP, they can still be used successfully in evaluating and improving on other LVLMs. It is interesting to see LLaVA's baseline model performing so strongly - we suspect this is because LLaVA is trained specifically for generating detailed descriptions, whereas InstructBLIP and mPLUG-OWL are more general models with a wide range of task applicability.

Additionally, we study the correlation between reward model and human evaluation scores. In Figure \ref{fig:reward_human_evaluation_correlation}, we see that across all human evaluated results, there is a clear and strong correlation between our reward model scores and human accuracy scores. Although this is by no means a robust replacement for human annotations, this shows the potential of training models as specific evaluators for hallucinations. Despite the noisiness, such a model could be used for early hyper-parameter selection, being much more cost-effective than humans evaluation.

\paragraph{Fine-Grained DPO}

We evaluate two variations of FDPO across the three classes - one that ignores analysis (IA), and one that disprefers analysis (DA), merging it with the inaccurate class. We see in Table \ref{tab:humanevalres} that marking analysis as a negative class does not impact hallucination rates in a significant way when training with FDPO, and may actually worsen rates at higher temperatures. We suspect that this may be because InstructBLIP's generations often have the last sentence being subjective analysis of the image, followed by an end of sequence token. Pushing down the likelihoods of generating this sentence increases the likelihood of the generation being lengthened, potentially inducing additional hallucinations as the model runs out of accurate content to describe.

On the other hand, we see that ignoring analysis in FDPO training almost cuts hallucination rates in half. Even sampling at high temperature, generations still on average contain less hallucinations than the baseline InstructBLIP model sampled at 0 temperature, where it would have the least propensity to hallucinate. This is slightly better than best-of-16 rejection sampling, and almost as good as best-of-64 rejection sampling. This performance gap is to be expected as rejection sampling can generalize over the entire set of possible model generations, whereas FDPO is more limited in optimizing only over the data that it sees in the training data. Though, there is a trade-off in this performance, as best-of-n rejection sampling is slower in inference by a factor of n. 

\section{Conclusion}

We introduce M-HalDetect, a novel multi-modal fine-grained hallucination detection dataset for benchmarking and training LVLMs to produce more truthful generations. We train fine-grained multi-modal reward models to perform rejection sampling against InstructBLIP. We innovate FDPO to optimize InstructBLIP directly on M-HalDetect, avoiding the need for preference pairs. Both methods significantly reduce InstructBLIP's hallucination rate, extending their effectiveness to the multi-modal domain, and demonstrating the usefulness of M-HalDetect in catching and reducing hallucinations. We show this dataset is generalizable across multiple LVLMs, successfully reducing the hallucination rates of LLaVA and mPLUG-OWL.

While we show strong performance with rejection sampling, it is prohibitively slow for inference in real-world use-cases. The next step would be to optimize a generative model, perhaps InstructBLIP,  using reinforcement learning with our trained reward models to create a higher quality LVLM for instruction aware VQA.

A limitation of modern day applications towards training large models with fine-grained feedback is that training typically takes place over multiple iterations of model training and feedback collection. This ensures the final model is more robustly aligned with the high level training objective. In this paper, we only perform one cycle of collecting response feedback and training. Indeed, when analyzing some of the responses, we can see hints of overfitting to our training objective - image descriptions are slightly more generic than before, and the preciseness of descriptions may have gone down. Future work can extend our dataset and methods to also account for descriptiveness and informativeness, training multiple reward models for optimizing a more robust final model. 
\section{Acknowledgements}
We thank Sean Hendryx and Utsav Garg for their feedback and support through internal development of the paper.

\bibliography{aaai24}
\clearpage
\end{document}